\documentclass[letterpaper]{article}
\usepackage{proceed2e}
\usepackage[margin=1in]{geometry}

\usepackage{times}
\usepackage{natbib}

\usepackage{amsmath}
\usepackage{mathrsfs}
\usepackage{color,soul}
\usepackage{bm}
\usepackage{dsfont}
\usepackage{graphicx}
\usepackage{multirow}
\usepackage{subcaption}
\usepackage{amssymb}
\usepackage{balance}

\usepackage{algorithm}
\usepackage{algorithmic}
\usepackage{capt-of}
\soulregister\cite7
\soulregister\citet7
\soulregister\citeA7
\soulregister\shortcite7
\soulregister\shortciteA7
\soulregister\citep7
\soulregister\ref7
\soulregister\pageref7
\soulregister\st7

\newcommand\bfu{\mathbf u}

\newcommand\bfx{\mathbf x}
\newcommand\bff{\mathbf f}
\newcommand\bfy{\mathbf y}

\newcommand\bfZ{\mathbf Z}
\newcommand\bfN{\mathbf N}
\newcommand\bfm{\mathbf m}
\newcommand\bfS{\mathbf S}
\newcommand\bfK{\mathbf K}

\newcommand\bfmu{\boldsymbol \mu}
\newcommand\bfSigma{\boldsymbol \Sigma}
\newcommand{\dee}{\,\textrm{d}}

\def\mycmd{0}

\if\mycmd1
\usepackage{tikz}
\usetikzlibrary{shapes, shapes.geometric, shapes.symbols, shapes.arrows, shapes.multipart, shapes.callouts, shapes.misc}
\usetikzlibrary{fit,positioning}
\usetikzlibrary{patterns}
\usepackage[utf8]{inputenc}
  \usepackage{fontspec}
  \usepackage{pgfplots}
  \usepgfplotslibrary{groupplots}
  \newlength\figurewidth
  \newlength\figureheight
  \pgfplotsset{compat=1.9}
  \usepgfplotslibrary{external}
  \tikzset{external/system call={lualatex
  \tikzexternalcheckshellescape -halt-on-error -interaction=batchmode -jobname "\image" "\texsource"}}
\fi

\title{Treatment-Response Models for Counterfactual Reasoning with Continuous-time, Continuous-valued Interventions}

\author{ {\bf Hossein Soleimani\thanks{ ~~These authors contributed equally to this work.}} \\
Computer Science Dept. \\
Johns Hopkins University\\
Baltimore, MD 21218 \\
\And
{\bf Adarsh Subbaswamy\footnotemark[1]}  \\
Computer Science Dept. \\
Johns Hopkins University\\
Baltimore, MD 21218 \\
\And
{\bf Suchi Saria}   \\
Computer Science Dept. \\
Johns Hopkins University\\
Baltimore, MD 21218 \\
}

\begin{document}

\maketitle

\begin{abstract}
Treatment effects can be estimated from observational data as the difference in potential outcomes. In this paper, we address the challenge of estimating the potential outcome when treatment-dose levels can vary continuously over time. Further, the outcome variable may not be measured at a regular frequency. Our proposed solution represents the treatment response curves using linear time-invariant dynamical systems---this provides a flexible means for modeling response over time to highly variable dose curves. Moreover, for multivariate data, the proposed method: uncovers shared structure in treatment response and the baseline across multiple markers; and, flexibly models challenging correlation structure both across and within signals over time. For this, we build upon the framework of multiple-output Gaussian Processes. On simulated and a challenging clinical dataset, we show significant gains in accuracy over state-of-the-art models.

\end{abstract}

\section{INTRODUCTION}\label{intro_section}

As computer storage is becoming cheaper, observational data from tracking user behavior are becoming increasingly available in many problem areas. In medicine, electronic health records contain data tracking a patient's disease progression over time. On the web, it is common for websites to track usage patterns over time. These data can be cheap and valuable sources for learning about the efficacy of interventions. For example, using health record data to study the effects of different drugs can accelerate our ability to generate hypotheses about drug responsiveness and design downstream experiments to discover unknown biology linked to efficacy \citep{wilke2011emerging}. Similarly, on the web, we can leverage observational data about visit and click patterns to measure user responses to interventions (e.g., a new ad campaign or a specific change to the website), allowing us to rapidly improve their web experiences; e.g., \cite{bottou2013counterfactual,li2015counterfactual,schnabel2016recommendations}.

In recent years, numerous studies have sought to use such non-experimental datasets for evaluating effects of interventions. For example, \cite{westreich2012parametric} use the parametric g-formula \citep{robins1987graphical} to estimate the effect of an antiretroviral therapy on time until AIDS or death from an observational cohort study. \cite{hong2008effects} use propensity score methods \citep{rosenbaum1984reducing} to examine the effect of kindergarten retention on self-perceived social health.

In this paper, we focus on the task of estimating effects in the setting where \textit{multiple} attributes (or outcome variables) are being measured over time. For example, in tracking kidney function, physicians measure several markers---creatinine, potassium levels, urine output---over time. The timing for when these measurements are made is driven by whether a patient's past  measurements suggest that they may be deteriorating. This means the markers are not sampled at regular intervals; rather, they may be missing at random \citep{rubin1976inference}. Further, there may be similarities in how individual markers change in response to treatment: for example, blood urea nitrogen (BUN) and creatinine, waste products in the blood, are filtered at similar rates during dialysis. Therefore, the treatment response to interventions for these markers are likely to be correlated.

\begin{figure}[!tbp]
\centering
\includegraphics[scale=0.53]{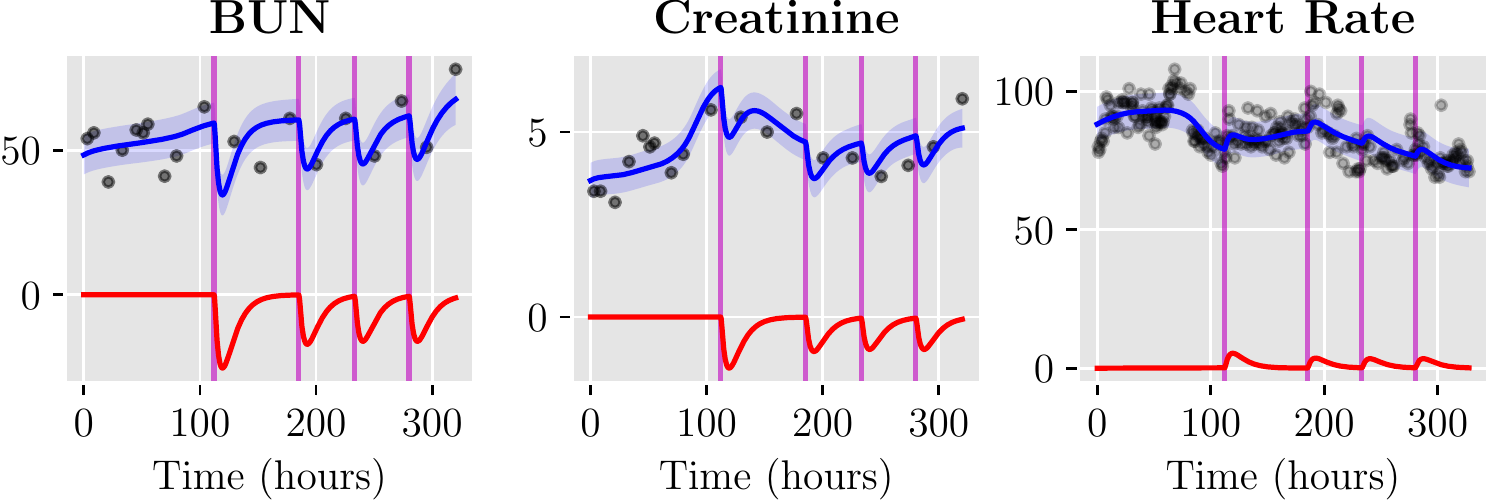}\vspace{-2mm}
\caption{The proposed model fits multivariate longitudinal data (blue) and extracts an estimated treatment response (red). Vertical purple lines denote the start of 3 hour dialysis sessions.}\vspace{-5mm} \label{structure_example}
\end{figure}

As an example, consider three signals from a patient, receiving the treatment of dialysis, shown in Fig. \ref{structure_example}. We show fits from the proposed model (blue) and its learned treatment response curves (red).
We see that the trajectories of BUN and creatinine are closely tied with similar responses (shown in red) to dialysis, while heart rate is less correlated with only a slight response to treatment.
We wish to infer such \emph{shared latent structure} in the treatment response curves from  multivariate longitudinal data. Doing so allows a better understanding of markers that may be coupled via a common latent process.

In causal inference, the problem of estimating the effect of the intervention (i.e. the treatment effect) has been studied extensively \citep{robins1987graphical,robins2000msm,gill2001causal,bang2005doubly}. Typically, the effect can be quantified as a difference in potential outcomes: the outcome under treatment and what the outcome would have been if no treatment were given \citep{rubin1974estimating,pearl2009causal}. Under specific assumptions about the data-generating process, the effect of the intervention is said to be identified and can be expressed as a function of the observational data. Once the effect has been determined to be identifiable, \emph{a key challenge is to posit an accurate and flexible estimation model for the outcome conditioned on history and treatment information}. In recent causal inference challenges, Bayesian additive regression trees (BART) \citep{chipman2010bart}, a Bayesian non-parametric method, have become an increasingly popular choice of model to wide success \citep{hill2011bayesian,green2012modeling}. However, popular outcome models from cross-sectional scenarios, such as BART, do not naturally extend to functional data.

Recently, \cite{xu2016bayesian} proposed a flexible Bayesian non-parametric approach for estimating \emph{univariate} treatment response curves and predicting disease trajectories. Their approach, which attempts to model the impact of a treatment over time, is limited to discrete-time treatments. This is consistent with most existing methods which model treatments as discrete events with potentially continuous doses  \citep{greenland1995dose,silva2016observational}. While some treatments are administered discretely (e.g., diuretics when given orally), in practice, many others, such as dialysis or intravenous diuretics, are administered continuously over a period of time. Naturally, these treatments should be modeled as continuous processes.

In this paper, we propose a semi-parametric Bayesian framework to model treatment effects in \emph{multivariate longitudinal data}. The proposed framework makes the following contributions. Contrary to past work, it unifies response modeling for both discrete and \emph{continuously-administered} treatments.
We build upon linear time-invariant (LTI) systems \citep{golnaraghi2010automatic} to flexibly represent the \emph{dynamic} response of physiologic signals to arbitrary treatments (treatment response curves). LTI systems and more generally differential equations are a natural and intuitive representation for describing the way in which an intervention causes change in the outcome over time. They are also used in pharmacokinetics for modeling concentration of drugs within the body over time \citep{cutler1978linear,shargel2005applied,rich2016optimal}.

Second, because data are sparse and irregularly sampled, rather than using imputation methods, we use multiple-output Gaussian Processes (GPs) to jointly model correlated multivariate signals through a shared low-dimensional latent function space. This has multiple advantages including flexibility to fit the complex variation seen within signals and across individuals in clinical data. However, GPs scale poorly with the number of observations. To circumvent this, we employ a fast, approximate inference algorithm using sparse variational inference techniques \citep{Titsias2009,Hensman2013,Hensman2015}.

The rest of the paper is organized as follows. We discuss related work in section \ref{relwork_section}. In section \ref{method_section}, we describe the proposed model. In section \ref{exp_section}, we present experimental results on a simulated and clinical dataset. Finally, concluding remarks are in section \ref{conc_section}.

\vspace{-2mm}
\section{RELATED WORK}\label{relwork_section}
\vspace{-3mm}

\paragraph{Continuous-Dose Treatment Modeling:}
Standard causal inference methods for estimating treatment effects primarily focus on discrete-valued, discrete-time interventions. Here, the effect of the intervention is typically estimated at a single point in time after the treatment is administered. For example, \cite{card1999causal} measures the effect of an individual's number of years of education (discrete-valued treatment) on their income.
Other studies have considered the effect of continuous-valued treatments. For instance, several studies have focused on estimating the effect of treatment on the outcome variable as a function of the treatment's continuous-valued dose; see, e.g., \citep{greenland1995dose, silva2016observational}. Others \citep{moodie2012estimation} estimate the effect of continuous-valued dose on multivariate longitudinal data using generalized propensity scores.

Most existing studies focus on the settings where the treatment is given at discrete time points. A few works which have considered continuously-administered treatments have been limited to special cases of dose-response learning. For example, \cite{johnson2005semiparametric} consider a problem in which treatment assignment is randomized, but the dose (the duration of continuous Integrilin therapy) is not. In this case, while the treatment is administered continuously, its duration is modeled as the dose and the effect is defined with respect to the outcome at a fixed moment in time.
\cite{tao2016semiparametric} also considers continuously-administered treatments in a similar way by modeling the time at which the treatment is discontinued as a time-to-event random variable for developing dynamic treatment regimes.
Both these approaches apply to the setting where the treatment dose is fixed upfront and does not vary within the duration of the treatment administration period.

\vspace{-3mm}
\paragraph{Modeling Irregularly-Sampled Multivariate Longitudinal Data:}
Generalized mixed-effects models \citep{verbeke2009linear} have been widely used to model multivariate longitudinal data. However, these models rely on strong parametric assumptions and thus cannot be easily applied to challenging data such as physiologic signals.

Several flexible probabilistic approaches have been proposed for modeling sparse and irregularly sampled longitudinal data.
Bayesian non-parametric models based on Gaussian processes (GPs) have been particularly effective in modeling physiologic time series. For example, \cite{Ghassemi2015} use multi-task GPs, \cite{schulam2016integrative} use coupled GPs, and \cite{Liu2016} combine state-space models with GPs.

Other flexible methods such as recurrent neural networks have also been used for modeling multivariate longitudinal data \citep{Lipton2016}. These methods require alignment of the measurements across signals. When observations are not aligned, imputation and other smoothing methods are used to fill in missing observations.

In this paper, we leverage multi-output GPs to jointly model multivariate physiologic signals.
In contrast to above-mentioned methods, we aim to model shared structure across signals and estimate dynamic response to both discrete-time and continuous-time treatments.


\vspace{-4mm}
\section{METHODS}\label{method_section}
\vspace{-4mm}

Here, we first review linear dynamic systems and describe their applicability to learning treatment response curves.
We then develop a semi-parametric approach for modeling {multivariate} longitudinal signals with shared structure and treatment response curves.

\vspace{-2mm}
\subsection{BACKGROUND: LTI SYSTEMS}\label{background_section}
\vspace{-2mm}

As we see in Fig. \ref{structure_example}, the observed values of BUN and creatinine (black dots) decrease in response to dialysis. However, this response is not instantaneous; rather, BUN and creatinine decrease slowly after the treatment is initiated. Similarly, after the treatment is discontinued, the effect does not disappear immediately; e.g., in Fig. \ref{structure_example}, the observed values of BUN and creatinine start to increase and eventually return to the elevated level prior to treatment. To capture this \emph{dynamic} behavior, we use linear time-invariant (LTI) systems.

\begin{figure}[t]
\centering
\includegraphics[scale=0.43]{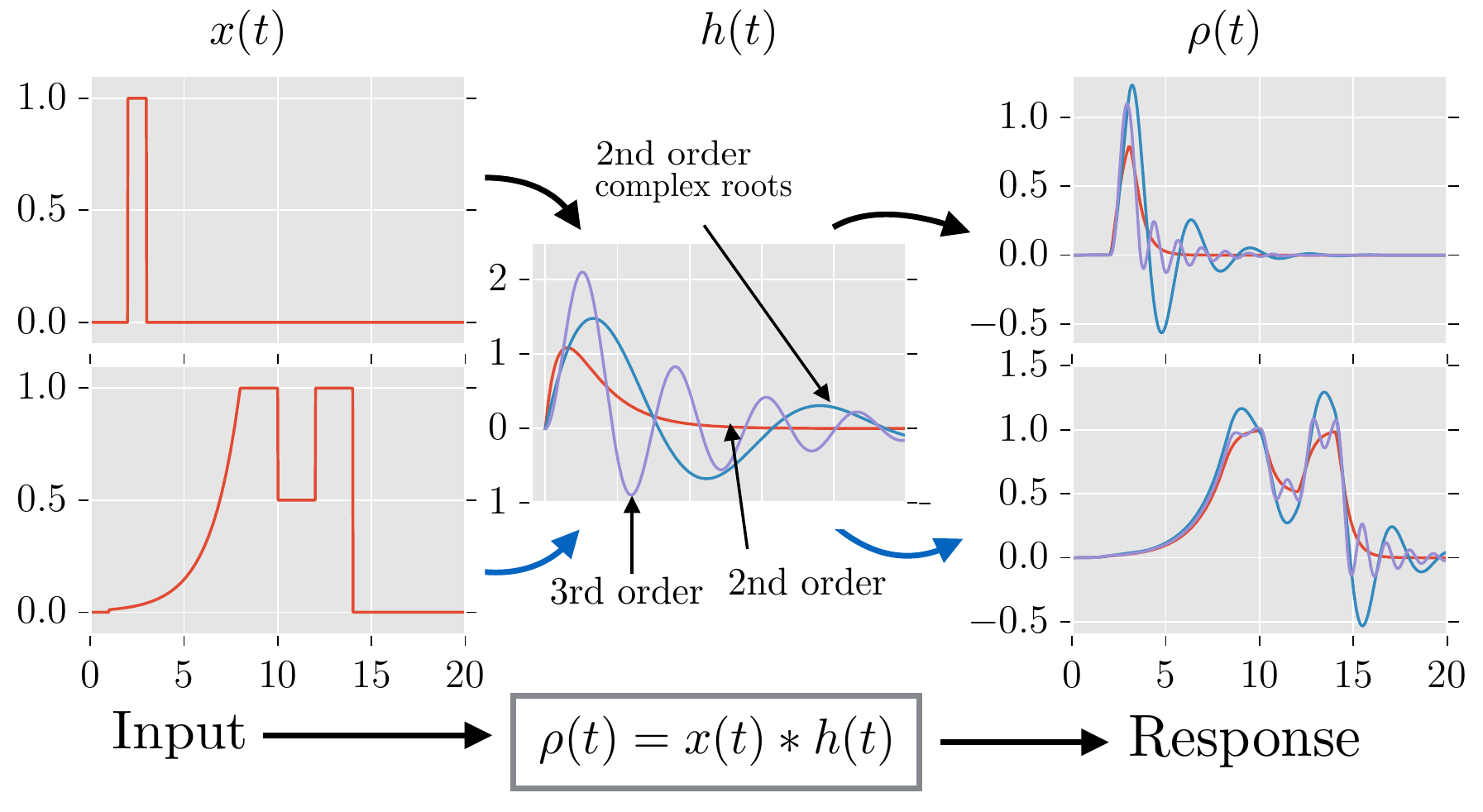}\vspace{-2mm}
\caption{Sample responses ($\rho(t)$) of three LTI systems, characterized by $h(t)$, to input signals ($x(t)$).}\vspace{-4mm} \label{sample_lti_fig}
\end{figure}

Consider the examples of LTI dynamic systems shown in Fig. \ref{sample_lti_fig}. Given an input (e.g., drug dose) $x(t)$, the treatment response function $\rho(t)$ is generated by convolving $x(t)$ with an impulse response function $h(t)$ \citep{golnaraghi2010automatic}:

\vspace{-8mm}
\begin{align}
\rho(t) = x(t) \ast h(t) = \int_{-\infty}^{\infty}x(\tau)h(t-\tau)\dee \tau\, .\label{conv_eqn}\vspace{-10mm}
\end{align}
LTI systems make the following two assumptions:

\noindent A1) Linearity: Suppose ($x_1(t), \rho_1(t)$) and ($x_2(t), \rho_2(t)$) are two pairs of input-output signals for the system. The response to the input $ax_1(t)+bx_2(t)$ is $a\rho_1(t)+b\rho_2(t), \forall a,b \in \mathbb{R}$. Thus, treatment responses are additive when multiple treatments are given in the same interval. \vspace{1mm}\\
\noindent A2) Time-Invariance: If $\rho(t)$ is the output of the system to input $x(t)$, the response to a shifted input signal $x(t-t_0)$ is $\rho(t-t_0)$ for every choice of $t_0\in \mathbb{R}$.

In the top and bottom left subplots of Fig. \ref{sample_lti_fig}, we show two different dose functions. The middle panel, shows impulse response functions, $h(t)$, for three LTI systems. Convolving the input (dose functions) with different impulse functions leads to varying treatment response curves as shown in the top and bottom right subplots.

The input-output relationship of LTI dynamical systems can also be characterized using differential equations. Consider, for instance, the differential equations for a second order LTI system $\frac{d^2\rho(t)}{dt^2} + (\alpha + \beta)\frac{d\rho(t)}{dt} + \alpha\beta = \alpha\beta x(t)$,
where $\alpha \neq \beta$ are positive constants. The impulse response for this system is:\vspace{-3mm}
\begin{align}
h(t) = \frac{\alpha\beta}{\beta-\alpha}(\mathrm{e}^{-\alpha t}-\mathrm{e}^{-\beta t})\mathds{1}(t\geq 0)\,.\vspace{-5mm} \label{impulse_eqn}
\end{align}\vspace{-0mm}
Here, the indicator function, $\mathds{1}(c)$, equals 1 or 0 when $c$ is true or false, respectively. The family of impulse response functions can be made more flexible by adding higher order derivatives to the differential equation.

\vspace{-1mm}
\subsection{PROPOSED MODEL}
In this section, we describe our proposed approach for modeling multiple outcomes and estimating dynamic treatment responses from multivariate longitudinal data. We model the disease trajectory of every signal $d$ for patient $i$ as follows:\vspace{-2mm}
\begin{align}
y_{id}(t) = \underbrace{\sum_{j=1}^J a_{ijd}(t;x^{0:t}_{ij})}_{\substack{\text{treatment response}}} + \underbrace{\phi_{id}(t)}_{\substack{\text{fixed-effects}\\\text{ component}}} + \underbrace{f_{id}(t)}_{\substack{\text{random-effects}\\\text{ component}}}+\epsilon_{id}(t)\,.\label{model}\vspace{-15mm}
\end{align}\vspace{-6mm}

Our model described in (\ref{model}) consists of three major components: treatment response, fixed-effects, and random-effects components. The treatment response component is the sum of responses to each treatment type $j=1,2,...,J$. We denote the response of signal $d$ to the treatment of type $j$ given the treatment input (dose) signal from time $0$ to $t$, $x^{0:t}_{ij}$, by $a_{ijd}(t;x^{0:t}_{ij})$. Example treatment inputs are shown in the left column of Fig. \ref{sample_lti_fig}.

The fixed-effects and random-effects components, which together we denote as the mixed-effects component, model the natural evolution of the signal independently of the treatments. For example, aging adults have higher baseline creatinine levels. Further, creatinine may drift upward when left untreated (e.g., patients with acute kidney injury). The fixed-effects term is specific to each signal, whereas the random-effects term captures the correlations across and within each signal.
Finally, we assume an additive Gaussian noise term $\epsilon_{id}(t) = \mathcal{N}(0, \sigma^2_{id}),\forall d=1,2,...,D,\forall t$.

We expand on each of these components in the next sections. Specifically, we describe the \textit{sharing} mechanism built into these components that allows us to jointly model all $D$ longitudinal signals for each individual.

\vspace{-2mm}
\subsubsection{Treatment Response Component}
We develop the treatment response component of our model based on the structure of second order LTI dynamic systems described in section \ref{background_section}.

Some treatments affect multiple signals in similar forms. For instance, creatinine and blood urea nitrogen (BUN), measures of kidney function, are similarly affected by dialysis. However, their responses differ from those of vital signs such as heart rate. To capture these differences, we posit that the total treatment response of signal $d$ is a mixture of two terms: a response shared across signals, $\rho^{(0)}$, and a response specific to $d$, $\rho^{(d)}$.

Specifically, we model the response of treatment type $j$ on signal $d$ as follows:\vspace{-2.5mm}
\begin{align}
a_{ijd}(t;x^{0:t}_{ij}) =& \chi_{ijd} \big(\psi_{ijd}\underbrace{\rho^{(0)}_{ij}(t; x^{0:t}_{ij})}_{\substack{\text{shared}}} \nonumber\\
&+ (1-\psi_{ijd})\underbrace{\rho^{(d)}_{ij}(t;x^{0:t}_{ij})}_{\substack{\text{signal-specific}}}\big),
\label{treatment_model}
\end{align}\vspace{-4mm}

Here, $\psi_{ijd} \in [0,1]$ controls the degree of mixing between the shared and signal-specific terms, while $\chi_{ijd} \in \mathbb{R}$ defines the direction and influences the magnitude of the response. To improve interpretability of the estimated treatment response terms, we place sparsity inducing priors on $\psi$ so that $a_{ijd}(t;x^{0:t}_{ij})$ is primarily determined by either the shared \emph{or} the signal-specific component.

We use second order LTI systems, introduced in section \ref{background_section}, as the form of the shared and signal-specific treatment response components. We compute each of these as the convolution of a treatment input signal which encodes dose and duration information, $x_{ij}^{0:t}$, and an impulse response function, $h_{ij}(t) = \frac{\alpha_{ij}\beta_{ij}}{\beta_{ij}-\alpha_{ij}}(\mathrm{e}^{-\alpha_{ij} t}-\mathrm{e}^{-\beta_{ij} t})\mathds{1}(t\geq 0)$. The signal-specific impulse response function, $h_{ijd}(t)$, has distinct parameters $\alpha_{ijd}, \beta_{ijd}$ for each signal $d$, while the shared-impulse response function, $h_{ijd}(t)$ only has one pair of parameters $\alpha_{ij0}, \beta_{ij0}$.

\paragraph{Priors:} Since the treatment response parameters are individual-specific, we posit a population-level prior on them. This enables sharing statistical strength across individuals and prevents over-fitting.

Specifically, we posit a Gaussian prior on the weighting coefficients: $\chi_{ijd}\sim\mathcal{N}(\bar{\chi}; 1), \forall i,j,d$. We also place log-normal priors on parameters of the treatment functions (since they have to be positive) $\alpha_{ijd}\sim \log\mathcal{N} (\bar{\alpha}, 1), \beta_{ijd}\sim \log\mathcal{N} (\bar{\beta}, 1), \forall d=0,1,...,D, \forall i,j$.
Further, we impose a Beta prior on the mixing coefficient $\psi$: $\psi_{ijd}\sim\text{Beta}(\lambda^{-1}_{\psi}, \lambda^{-1}_{\psi})$.

\subsubsection{Fixed-Effects Component}
To capture covariate-dependent general trends in the progressions of particular markers (e.g., rate of deterioration can depend on age), we define the fixed-effects term, $\phi_{id}(t)=\bm{\gamma}^T_{id}\bm{c}_{it}$ as a deterministic linear regression term. Here, $\bm{c}_{it}$ is a vector of individual-specific covariates including observation time, and $\bm{\gamma}_{id}$ is the vector of regression parameters.
We place Gaussian priors on the parameters of the baseline regression: $\bm{\gamma}_{id} \sim \mathcal{N}(\bm{\bar{\gamma}}, I), \forall i,d$.

\subsubsection{Random-Effects Component}
The random-effects component captures correlations within and across signals.
Since physiologic signals typically have very challenging structure, which is difficult to capture using simple parametric functions, we use Gaussian processes as the building block of the random-effects component.

In particular, the random-effects component is comprised of a latent function shared across signals, $g_{i}(t)\sim \mathcal{GP}$, and a signal-specific latent function for every $d$, $v_{id}(t)\sim\mathcal{GP}$.
Some physiologic signals are expected to be strongly correlated;  e.g., BUN and creatinine, or heart rate and blood pressure. The shared latent component captures the common structure and correlations across signals. Further, each marker may have its own unique structure which cannot be modeled by $g_i(t)$; this signal-specific structure is modeled by $v_{id}(t)$.

The structure used in the random-effects component is similar to the linear models of coregionalization \citep{Teh2005,Alvarez2009}, which is used to jointly model multiple correlated signals.

Specifically, we define the random-effects component for each signal $d$ of every individual using a shared and a signal-specific term:
\begin{align}
f_{id}(t) = \underbrace{\omega_{id} g_{i}(t)}_{\substack{\text{shared component}}} + \underbrace{\kappa_{id} v_{id}(t)\,.}_{\substack{\text{signal-specific component}}} \label{baseline_model}
\end{align}\vspace{-5mm}

Here, $\omega_{id}, \kappa_{id}, \forall i,d,$ are the mixing coefficients, and $g_{i}(t)$, is a shared latent function drawn from a GP prior with zero mean and kernel function  $K_{i}$; i.e., $\mathbf{g}_{i} = g_{i}(\mathbf{t}_{id})\sim \mathcal{GP}(\mathbf{0}, \mathbf{K}^{(i)}_{N_{id}N_{id}})$ where $\mathbf{K}^{(i)}_{N_{id}N_{id}} = K_{i}(\mathbf{t}_{id}, \mathbf{t'}_{id})$, and $N_{id}$ is the number of observations from signal $d$. The signal-specific component is another latent function with a GP prior: $\mathbf{v}_{id} = v_{id}(\mathbf{t}_{id})\sim \mathcal{GP}(\mathbf{0}, \mathbf{K}^{(id)}_{N_{id}N_{id}})$ where $\mathbf{K}^{(id)}_{N_{id}N_{id}} = K_{id}(\mathbf{t}_{id}, \mathbf{t'}_{id})$. In contrast to the shared latent function, the latent functions $\mathbf{v}_{id}$ are drawn independently from a GP with a signal-specific kernel function.

We use Mat\'ern-3/2 kernel for each latent function (see, e.g., \citet{Rasmussen2006}) with variance set to 1 and length-scale as the only free parameter ($l^{(g)}_{i}$ for shared and $l^{(v)}_{id}$ for signal-specific latent functions). The variance is set to 1 since we can scale each latent function using the mixing coefficients $\omega$ and $\kappa$.

\paragraph{Priors:} We place Gaussian priors on the mixing coefficients: $\omega_{id}\sim\mathcal{N}(\bar{\omega}_{d}, \lambda^{-1}_{\omega\kappa})$ and  $\kappa_{id}\sim\mathcal{N}(\bar{\kappa}_{d}, \lambda^{-1}_{\omega\kappa})$, and log-normal priors on the length-scales of the kernels: $l^{(g)}_{i}\sim\log\mathcal{N}(\bar{l}^{g}, 1), \forall i$ and $l^{(v)}_{id}\sim\log\mathcal{N}(\bar{l}^{v}_{d}, 1), \forall i$.

\subsubsection{Learning and Inference}\label{approx_inf_section}
In this section, we describe the learning and inference for our proposed model. Our model has \textit{local} and \textit{global} parameters. Local parameters are individual-specific which we denote by $\Theta_i = \{\chi_{ijd}, \psi_{ijd}, \alpha_{ijd}, \beta_{ijd}, \alpha_{ij0}, \beta_{ij0}, \omega_{id}, \kappa_{id}, l^{(g)}_{i}, l^{(v)}_{id}, \sigma^2_{id}\}$. The global parameters, denoted by $\Theta_0$, are the parameters of the prior distributions which are shared across individuals; $\Theta_0=\{\bar{\chi}, \bar{\alpha}, \bar{\beta}, \bar{\gamma}, \bar{\omega}_{d}, \bar{\kappa}_{d}, \bar{l}^{g}, \bar{l}^{v}_{d}, \forall d\}$.

We take a maximum a posteriori (MAP) approach and compute point-estimates of all model parameters. We treat $\lambda_{\omega\kappa}$ and $\lambda_\psi$ as regularization terms. 
For each individual $i$, we observe the  longitudinal samples and the treatment inputs.
Computing the log-likelihood for each individual requires integrating out all latent stochastic function $g_{i}$ and $v_{id}$ in the random effects component (\ref{baseline_model}).

The main bottleneck for learning and inference in our model is the use of GPs in $f_{id}(t)$. Due to requiring covariance matrix inversion, GP inference scales cubically in the number of observations. To reduce this computational complexity, we develop the learning and inference algorithms for our model using the sparse variational techniques \citep{Titsias2009,Hensman2013,Hensman2015}.

Using these techniques, we compute the evidence lower bound for each individual: $\text{ELBO}_i$. Detailed derivation of $\text{ELBO}$ is provided in Appendix \ref{append_section}.

\paragraph{Learning:}
The overall objective function for our model (ELBO) is additive over the lower bound for each of the $I$ individuals: $\text{ELBO} = \sum_{i=1}^{I}\text{ELBO}_i$. We use stochastic gradient techniques for learning. At each iteration of the algorithm, we randomly select a mini-batch of individuals and update their local parameters, keeping the global parameters fixed. We then compute the gradients of ELBO on the mini-batch with respect to $\Theta_0$ and perform one step of stochastic gradient ascent to update the global parameters. We use AdaGrad \citep{Byrd1995} for stochastic gradient optimization. We repeat this process until either the relative change in global parameters is less than a threshold or the maximum number of iterations is reached.

We use TensorFlow \citep{Abadi2016} and GPflow \citep{gpflow} to implement our model. TensorFlow automatically computes gradients of the objective function with respect to model parameters.

\vspace{-2mm}
\section{EXPERIMENTS}\label{exp_section}\vspace{-2mm}
In this section, we evaluate the proposed model using two datasets. First, we perform a simulation study to show that in the presence of known shared structure (in signals and treatment effects) the proposed model can recover the true decomposition. In order to do this, we construct a synthetic dataset that mimics the conditions we expect to see in clinical data. Then, to demonstrate that the model can accurately represent complex multivariate longitudinal data, we compare the model against state-of-the-art baselines on the task of predicting disease trajectories under treatment in a challenging, publicly available clinical dataset. We also assess the use of the model as an exploratory tool for discovering shared structure in data of this sort by validating the learned treatment responses against clinical knowledge.

\begin{figure}[t!]
\centering
    \includegraphics[scale=0.53]{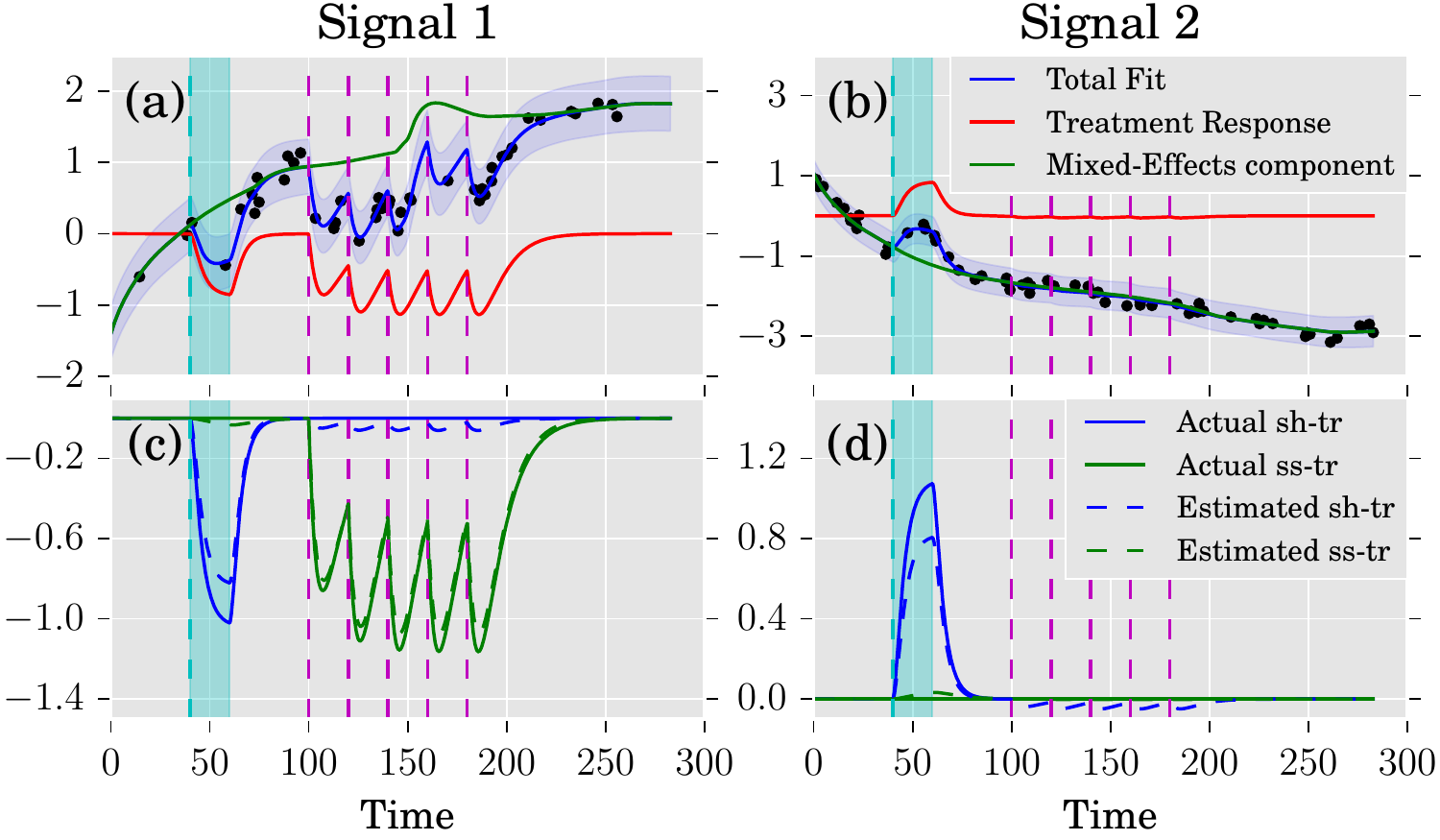}\vspace{-3mm}
\caption{Decomposition of the fit for the disease trajectory of one synthetic patient. (a)-(b) show the observations (dots) and the total fit's (blue curve) separation into mixed-effects (green curve) and treatment response (red curve) components for each signal. (c)-(d) show the estimated shared (sh-tr) and signal-specific (ss-tr) treatment response components against the true treatment response curves used to generate the data.} \label{synthetic_data_fig}\vspace{-5mm}
\end{figure}
\vspace{-2mm}
\subsection{SIMULATION STUDY}
\vspace{-2mm}
\paragraph{Data:} We generate the synthetic dataset consistent with the settings typically encountered in clinical observational studies. We generated sparse and irregularly sampled observations for each marker; observation times are drawn from a Poisson process. Specifically, we generated a synthetic dataset consisting of 50 patients, each with two \textit{irregularly-sampled longitudinal signals}. To simulate the unalignment of the signals, for every marker of each patient, we first randomly chose the number of observations from a Poisson distribution with mean 50, and then generated the observation times from a Poisson process with rate 0.2.

We assume each signal consists of a mixed-effects and a treatment response component.
We generated the signals with \emph{shared mixed-effects} components. In particular, the signals have opposite fixed-effect trends with similar rates and shared random-effect component drawn from a GP. For each patient $i$, we computed the mixed-effects component using $\gamma_{id1} \log(t+10) + \gamma_{id2} + f_i(t)$, where $\gamma_{id1},\gamma_{id2}$ are signal-specific coefficients and $f_i\sim\mathcal{GP}(0,K)$ with a shared patient-specific RBF kernel with length-scale 20 and variance 0.05. We assume patients in the population have similar fixed-effect components; we draw $\gamma_{i1}\sim \mathcal{N}(1, 0.1), \gamma_{i2}\sim \mathcal{N}(-4, 0.1)$ for signal 1 and $\gamma_{i1}\sim \mathcal{N}(-1, 0.1), \gamma_{i2}\sim \mathcal{N}(4, 0.1),\forall i$ for signal 2.

In clinical data, some treatments affect multiple physiologic markers whereas others may only change one marker. To simulate this setting, we consider two treatments, a continuous one which has shared effect on both signals and a discrete treatment with signal-specific effect only on the first marker.
We generate treatment 1 with $\alpha_{i1}\sim\log\mathcal{N}(0.2, 0.1), \beta_{i1}\sim\log\mathcal{N}(0.6, 0.1)$, and treatment type 2 with $\alpha_{i2}\sim\log\mathcal{N}(0.1, 0.1), \beta_{i2}\sim\log\mathcal{N}(0.2, 0.1)$. Treatment type 1 has shared effects on both signals, with $\chi_{i11}\sim\mathcal{N}(-1, 0.1)$ and $\chi_{i12}\sim\mathcal{N}(1, 0.1)$. This treatment was administered continuously in the time interval $[40,60]$ with dose 1.
Treatment 2, on the other hand, only has signal-specific effect on marker 1 with $\chi_{i21}\sim\mathcal{N}(1, 0.05)$ and $\chi_{i22}=0$, and was given discretely at times 100, 120, 140, 160, and 180, each with dose 20.

We train our model on the trajectories from all 50 patients, setting the maximum number of iterations for optimization to 1000 with mini-batch size 2.

\begin{figure}[t!]
\centering
    \includegraphics[scale=0.53]{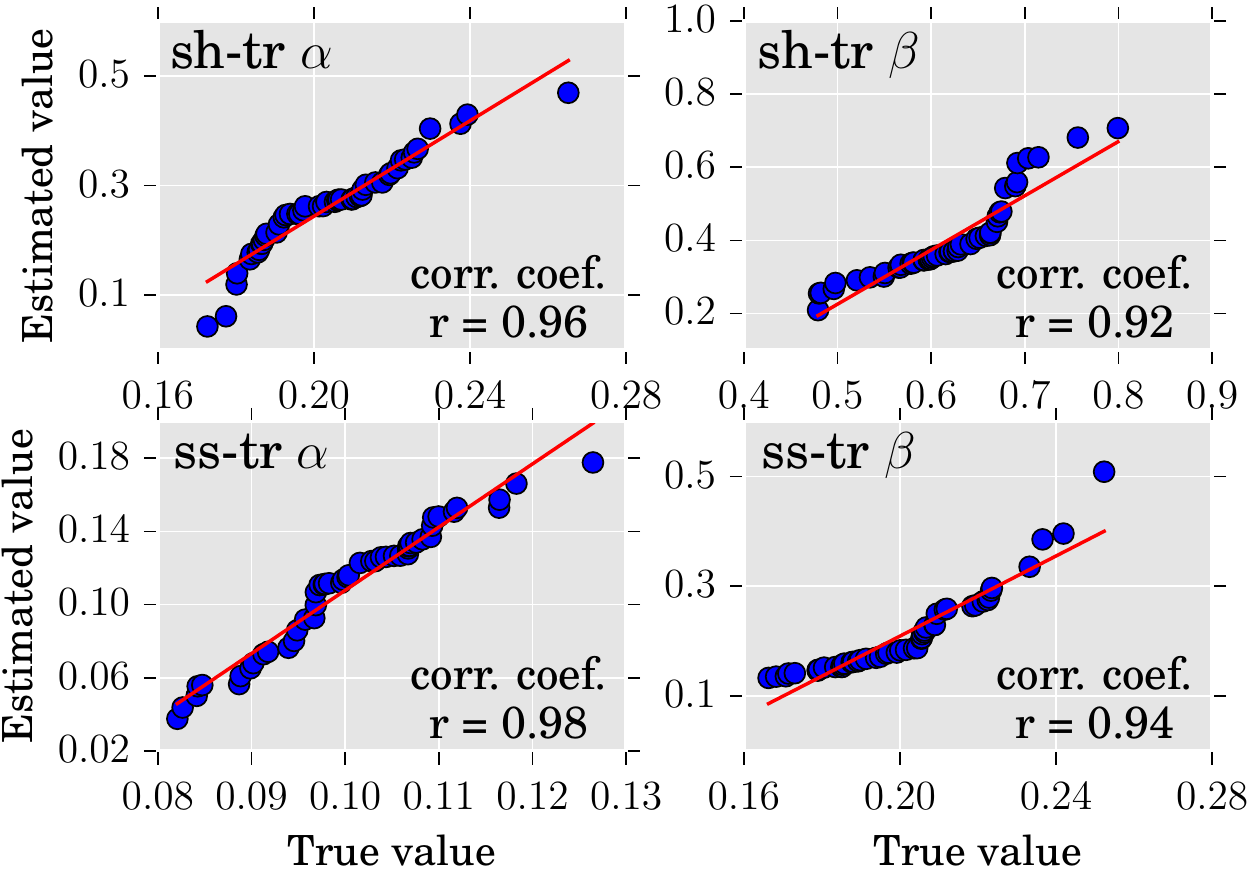}\vspace{-3mm}
\caption{Q-Q plot for the estimated and the true $\alpha$, $\beta$ of shared and signal-specific treatment response functions.} \vspace{-6mm}\label{qq_fig}
\end{figure}%

\vspace{-4mm}
\paragraph{Results:} To evaluate the ability of the proposed model to fit disease trajectories, in Fig. \ref{synthetic_data_fig} we show the decomposition of the model's fit for one patient. Fig. \ref{synthetic_data_fig}a-b shows, for each of the signals, the observations (dots) and the total fit's (blue curve) separation into mixed-effects (green curve) and treatment response (red curve) components. The purple vertical lines correspond to treatment type 2, and the shaded cyan regions indicate the duration of treatment type 1. These plots qualitatively show that the estimated signal values closely fit the disease trajectory under the given treatments. Note that the proposed model is able to uncover that treatment type 2 only affects signal 1, and, as a result, learns that there are no shared effects. Rather, the effect on signal 1 is correctly captured using the signal-specific component.

To analyze the correctness of the decomposition, in Fig. \ref{synthetic_data_fig}c-d we plot the estimated (dashed lines) shared and signal-specific treatment response components against the true (solid lines) treatment response curves used to generate the data. We can see that the model does learn the correct treatment responses. Further, we compare the true and estimated $\alpha, \beta$ parameters of the treatment functions for all patients using the Q-Q plots in Fig. \ref{qq_fig}. This figure shows that our model can closely recover the true $\alpha$ and $\beta$ since the relationship between the quantiles of the distributions of the estimated and true values is linear (with correlation coefficients greater than 0.92).

\vspace{-2mm}
\subsection{REAL DATA: MODELING RESPONSE TO DIALYSIS}
\vspace{-1mm}

\paragraph{Dataset:}
We apply and evaluate our model on the MIMIC II Clinical Database \citep{saeed2002mimic}, which consists of electronic health records for ICU patients from the Beth Israel Deaconess Medical Center. In particular, we examine the effect of two types of dialysis: Continuous Renal Replacement Therapy (CRRT) and Intermittent Hemodialysis (IHD).

Dialysis is used to filter blood in place of the kidneys when patients are suffering from acute kidney injury (AKI). While CRRT and IHD are both types of hemodialysis, they differ in their administration: IHD is typically administered 3 times a week with each session lasting 3 hours, whereas CRRT must be administered in the ICU and is typically given 24 hours a day \citep{pannu2005renal}. In the ICU, CRRT is used as an alternative for IHD with no significant differences in effect \citep{pannu2005renal}.

Elevated levels of waste products like blood urea nitrogen (BUN) and creatinine, which indicate poor kidney function, are reduced over the course of dialysis. Similarly, dialysis is used to regulate the concentration of electrolytes like potassium and calcium in the blood. Dialysis sometimes also indirectly affects other signals which are not primarily controlled by the kidneys; e.g., it can reduce blood pressure causing hypotension \citep{chou2006physiological}.
Therefore, we chose to model BUN and creatinine as primary measures of kidney function, potassium and calcium since they are dialysis-regulated solutes in blood, and blood pressure and heart rate as signals which may be affected by dialysis but are typically not primary factors of interest when prescribing dialysis.

We included acute kidney injury patients who had at least 10 measured values in each signal and received no treatments other than dialysis for AKI, resulting in 67 relevant patients.
For every patient, we use the first 70\% of the observed marker trajectories for training and the rest for prediction.
On average, each patient has $106$ training observations and $46$ test observations per signal. The training and test regions have average lengths of $11$ and $6$ days, respectively.

The learning algorithm for the proposed model alternates between optimization of local and global variables. We set the learning rate and the maximum number of iterations for the global optimization step to $0.05$ and $200$, respectively. The local optimization for each individual is terminated when the relative change in the objective function $(\text{ELBO}_i)$ is less than $10^{-4}$ or the maximum number of iterations ($500$) is reached. We also set the regularization terms $\lambda_{\psi}=100$ and $\lambda_{\omega\kappa}=0.1$. These were determined on held-out training data from a subset of the patients. We found that the model performance was not highly sensitive to the choice of the regularization terms within a reasonable range. \vspace{-1mm}

\begin{figure}[!tb]
\centering
\includegraphics[scale=0.55]{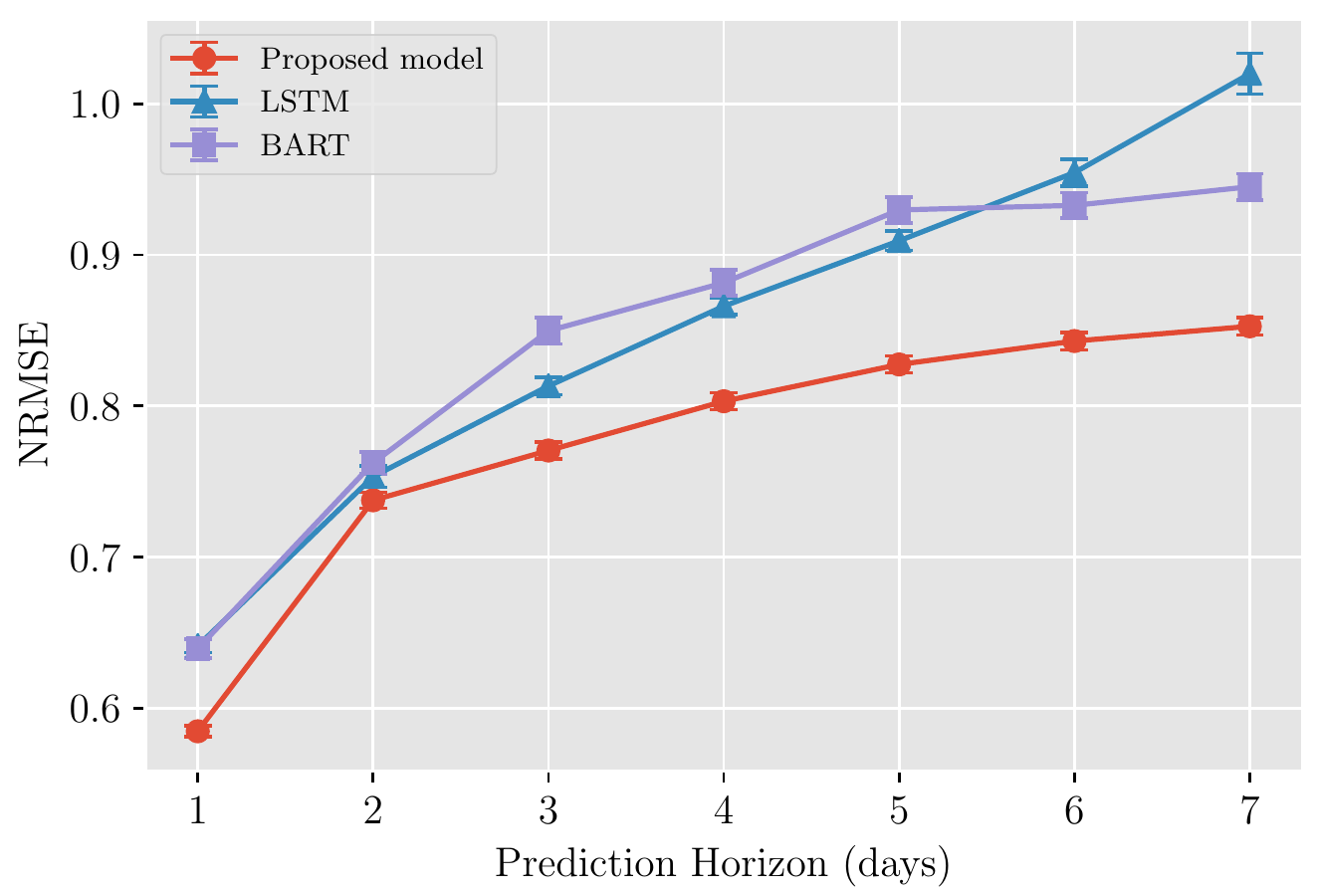}\vspace{-2mm}
\caption{Normalized root mean square error (NRMSE) for the proposed model and the baselines.} \vspace{-4mm}\label{rmse_example}
\end{figure}

\vspace{-2mm}
\paragraph{Baselines:} We compare the proposed model's ability to predict marker trajectories against two state-of-the-art regression and time series prediction methods: Bayesian additive regression trees (BART) \citep{chipman2010bart}, and long short-term memory (LSTM) \citep{hochreiter1997long}.

For both baselines, we train separate models for each signal by binning the data into 6 hour intervals. For non-empty intervals, we take the average value. When there are no observations in an interval, we take the value of the previous bin; i.e., we use last-observation-carried-forward imputation. We use the bin midpoint, time since last treatment, dose of last treatment, and the marker value as features.  We train BART and LSTM with the features from the previous $L$ bins as covariates and the current value of the marker as the outcome.
We determine $L$ for each method via cross-validation. These methods are trained for one-step ahead prediction. For predictions longer than one step, we use the model's predicted value as a feature for predicting the next step.

\begin{figure*}[t]
\centering
\includegraphics[scale=0.53]{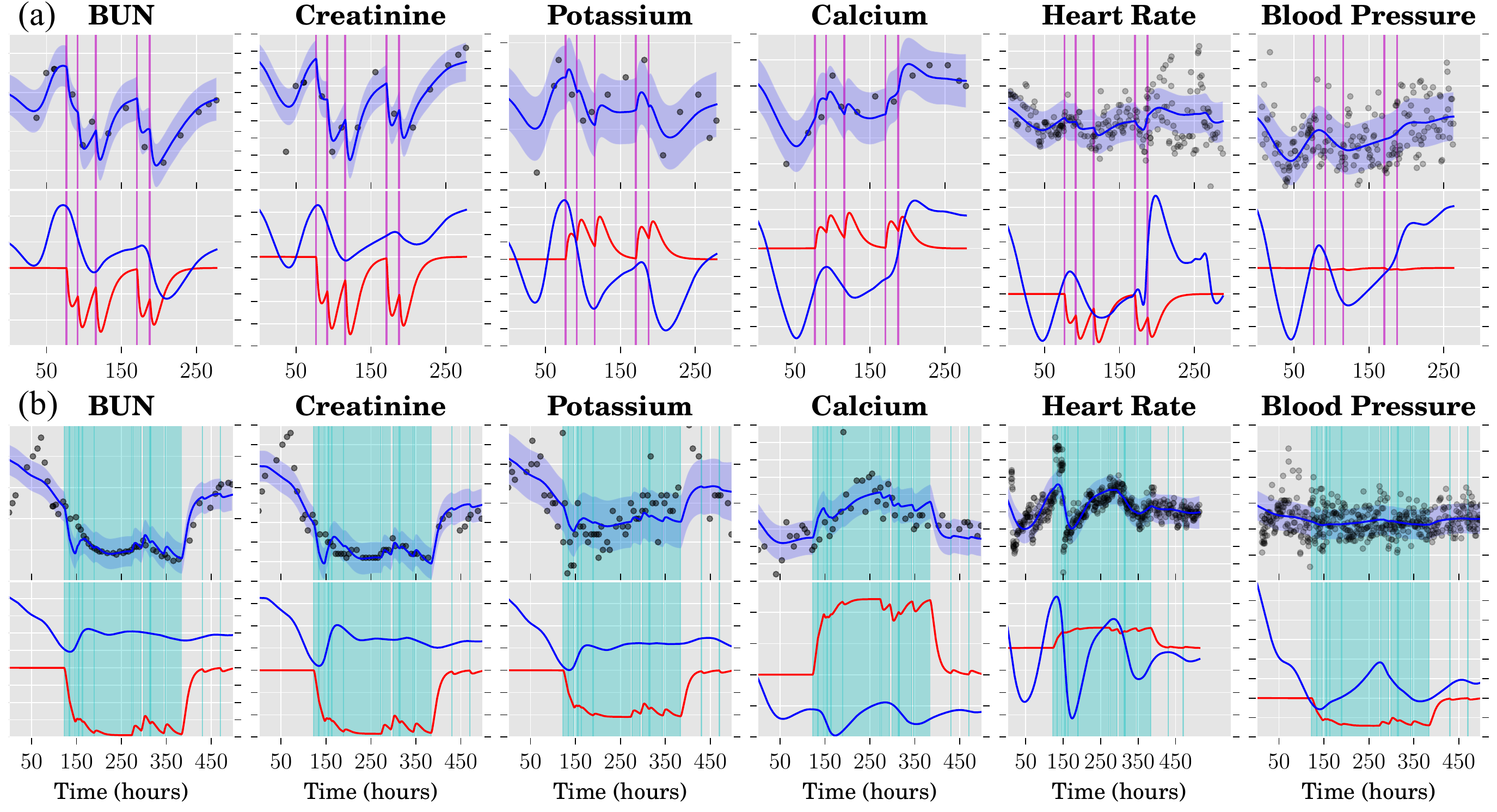}\vspace{-2mm}
\caption{The observations (dots), the model's learned fits (blue line, top row) and mixed-effects (blue line, bottom row, shifted down for viewing) and treatment response (red line, bottom row) decompositions for 6 marker trajectories of two patients receiving (a) IHD (purple vertical lines) (b) CRRT (cyan shaded regions).} \vspace{-5mm}\label{qualitative}
\end{figure*}

\vspace{-3mm}
\paragraph{Performance Criterion:} We compute the root mean square error normalized by the standard deviation for each signal (NRMSE) and then average across markers. We also performed non-parametric bootstrap with sample size 50 on the test set to measure the statistical significance of the results.

\paragraph{Quantitative Results:}
Fig. \ref{rmse_example} shows NRMSE and standard error for prediction horizons of 1 to 7 days for each method.
We see that the proposed method achieves better NRMSE than BART and LSTM. Further, the rate of increase of NRMSE as a function of prediction horizon is smaller for the proposed method than the baselines. Neither BART nor LSTM can naturally model treatment response. This is especially apparent on the subset of patients who received treatment in the test region: for a prediction horizon of 7 days on these patients, the model does $15\%$  better than BART and $8\%$ better than LSTM.

\paragraph{Qualitative Results:} In order to demonstrate the potential of our model as an exploratory analysis tool, we examine in detail the latent treatment response structure it uncovers on the kidney function data.

Fig. \ref{qualitative}a,b show the model's learned fits (top row) and mixed-effects and treatment response decompositions (bottom row) for six physiologic trajectories of two patients, a patient receiving IHD (Fig \ref{qualitative}a) and a patient receiving CRRT (Fig. \ref{qualitative}b). We can see that the proposed model can fit the overall trajectories in these signals well.

The learned decompositions qualitatively match clinical knowledge of dialysis. Both types of dialysis are used to lower BUN and creatinine levels in the blood. This is seen in both patients, for whom the proposed model learns a negative treatment response curve (red line in the bottom row). Further, we see that the observed values of BUN and creatinine return to elevated values once treatment is discontinued, as would be clinically relevant.

In order to quantitatively compare the relative impact of each treatment type $j$, we define a criterion to compute its maximum effect on every signal $d$, ($\mathcal{I}_{jd}$). Since the 2nd order LTI systems used in the treatment response components are normalized, the maximum amplitude of the estimated treatment response curves is bounded by the volume (dose) of the treatment input signal, $x(t)$, multiplied by the corresponding coefficient $\chi$. Thus, we compute $\mathcal{I}_{ijd} = \chi_{ijd} \times [\max_t x_{ij}(t)]$ for each individual $i$.

Since dialysis is primarily used for managing waste products like BUN and creatinine, and electrolytes like calcium and potassium, the estimated effect of CRRT and IHD on these signals should be higher than the estimated effects on vital signals. This matches what our model learns: on average across the population, the maximum treatment effect of dialysis ($\mathcal{I}$) on BUN (-5.14), creatinine (-3.39), potassium (-0.57), and calcium (2.03) is greater than $\mathcal{I}$ on heart rate (1.48) and blood pressure (-0.08), which can be indirectly affected by dialysis.

As we qualitatively saw in Fig. \ref{qualitative}, dialysis decreases BUN and creatinine. This is consistent with the average $\mathcal{I}$ of these two signals which are both negative. We find that the maximum effect of BUN is significantly less than 0 using a one-sided t-test with p-value $0.015$, though it is not significant for the max effect of creatinine at the $0.05$-level with p-value $0.065$.

While CRRT and IHD are expected to have similar outcomes with respect to BUN and creatinine, IHD sessions are shorter with more aggressive dosing. The proposed model captures this difference: $\mathcal{I}_{\text{BUN}}$ and $\mathcal{I}_{\text{creatinine}}$ are significantly lower for CRRT than for IHD with p-values $0.0002$ and $0.013$, calculated using a one-sided t-test.

The ability of the model to extract these clinically-validated relationships from the sparse, unaligned, multivariate data shows its capability as an exploratory tool.
\vspace{-2mm}
\section{CONCLUSION}\label{conc_section}
\vspace{-2mm}
In this paper, we propose a flexible Bayesian semi-parametric approach for modeling multivariate outcomes over time. The proposed method provides a unified way to model continuous response over time to highly flexible dose functions (i.e. functions with continuous changes in dose level over time). For this, we model the treatment response curves using linear time-invariant dynamical systems. Further, our approach enables learning of structure in the response function that may be shared across multiple signals. Finally, using both simulated and real datasets, we show significant gains in performance for predicting outcomes.

It is worth noting that in the potential outcomes framework, estimation of the causal effect is precluded by the need to justify a set of generally untestable assumptions \citep{gill2001causal}; e.g., see \cite{lok2008statistical,schulam2017whatif} for assumptions needed for continuous-time potential outcomes. When these conditions hold, the proposed method provides a highly flexible and accurate means for jointly modeling the multivariate potential outcomes. Further, while our approach relies on regularization to decompose the observed data into shared and signal-specific components, we need new methods for constraining the model in order to guarantee posterior consistency of the sub-components of this model.

\subsubsection*{Acknowledgements}
The authors would like to thank Yanbo Xu for valuable discussions.

\vspace{-2mm}

\newpage
\appendix
\vspace{-3mm}
\section{Appendix}\label{append_section}
\vspace{-3mm}

We observe $N_{id}$ samples from each signal $d$ of every individual $i$; $\mathbf{y}_{id} = y_{id}(\mathbf{t}_{id}) = \{y_{id}(t_{idn}),\forall n=1,2,...,N_{id}\}$. We denote the collection of observations of $D$ longitudinal signals by $\mathbf{y}_i = \{\mathbf{y}_{i1},...,\mathbf{y}_{iD}\}, \forall i$. We also observe the treatments given to each individual $i$, $\bfx_{ijd}(t), \forall t, j, d$. We let $\mathbf{x}_i = \{\mathbf{x}_{i1},...,\mathbf{x}_{iJ}\}, \forall i,$ be the collection of treatment inputs of all types.

For each latent function, we define a set of inducing input-output pairs $\bfZ, \bfu$, where $\bfZ$ are some pseudo-inputs, known as inducing points, and $\bfu$ are the values of the Gaussian process at $\bfZ$. We place the inducing points $\bfZ$ on a grid. We define a variational distribution for $\bfu$, $q(\bfu) = \mathcal{GP}(\bfm, \bfS)$, where $\bfm$ and $\bfS$ are variational parameters. Using $q(\bfu)$ we compute a variational GP distribution for each shared latent function: $q(\mathbf g) = \mathcal{GP}(\bfmu_{g}, \bfSigma_{g})$, where $\bfmu_{g} = \bfK_{\bfN \bfZ}\bfK^{{-1}}_{\bfZ\bfZ}\bfm$ and $\bfSigma_{g} = \bfK_{\bfN\bfN}-\bfK_{\bfN\bfZ}\bfK^{{-1}}_{\bfZ\bfZ}(\mathbf I - \bfS\bfK^{{-1}}_{\bfZ\bfZ})\bfK_{\bfZ\bfN}$, with $\bfK_{\bfN\bfZ} = K(\mathbf{t}, \bfZ)$. We similarly obtain $q(\mathbf v_{d}) = \mathcal{GP}(\bfmu_{v_{d}}, \bfSigma_{v_{d}}), \forall d,$ for signal-specific latent functions.

Here, to simplify the notation, we assume $\mathbf t_{id} = \mathbf{t}_i, \forall d,$ and write $\bfK_{\bfN\bfN} = K_{i}(\mathbf{t}_{i}, \mathbf{t}'_{i})$. We emphasize that the observations from different signals need not be aligned for our learning and inference algorithm.

We obtain the variational distribution $q(\bff)$ by taking the linear combinations of the variational distributions for individual latent GPs: $q(\bff_d) = \mathcal{GP}(\bfmu_{d}, \bfSigma_d)$, where $\bfmu_{d} = \omega_{d}\bfmu_{g} + \kappa_d\bfmu_{v_d}$ and $\bfSigma_{d} = \omega^2_{d}\bfSigma_{g} + \kappa^2_d\bfSigma_{v_d}$. 

The log-likelihood of the observations and local model parameters for each individual is $\log p(\bfy_i,\Theta_i)=\log p(\bfy_i|\Theta_i) + \log p(\Theta_i)$, where we dropped the explicit conditioning on $\bfx_i$. Using sparse GP approximations and Jensen's inequality, we compute a variational lower bound for $\log p(\bfy_i|\Theta_i)$:
\begin{align}
\log p(\bfy_i|\Theta_i) &= \log\int p(\bfy_i| \bff_i)p(\bff_i|\bfu_i)p(\bfu_i)\dee \bff_i\dee \bfu_i\nonumber\\
&\leq E_{q(\bff_i)}\log p(\bfy_i|\bff_i, \Theta_i) - \text{KL}(q(\bfu_i)||p(\bfu_i))\nonumber\\
& =Q_i(\bfy_i;\Theta_i)\, ,  \label{var_lower_bnd}
\end{align}
where, $q(\bff_i) = E_{q(\bfu_i)}p(\bff_i|\bfu_i)$. Also, $\text{KL}(q(\bfu_i)||p(\bfu_i))$ is the Kullback-Leibler divergence between $q(\bfu_i)$ and $p(\bfu_i)$ which we compute analytically.
We note that conditioned on $\bff_i$, the distribution of $\bfy_i$ factorizes over all signals. 
Thus, we have $E_{q(\bff_i)}\log p(\bfy_i|\bff_i, \Theta_i) = \sum_{d} [\log {m}_{id}(\mathbf{t}_{id}) + E_{q(\bff_{id})}\log p(\bfy_{id}|\bff_{id})]$, where $m_{id}$ is the sum of the treatment response component and the fixed-effect terms. The expectation $E_{q(\bff_{id})}\log p(\bfy_{id}|\bff_{id})$ is also available in closed-form \citep{Titsias2009,Hensman2013}.
Given (\ref{var_lower_bnd}), we compute the evidence lower bound (ELBO) for each individual: 
$\text{ELBO}_i = Q_i(\bfy_i;\Theta_i) + \log p(\Theta_i)$.

\end{document}